\documentclass[journal]{IEEEtran}
\usepackage{cite}
\usepackage{amsmath,amssymb,amsfonts}
\usepackage{graphicx}
\usepackage{textcomp}
\usepackage{siunitx}
\usepackage{algorithmic}
\usepackage{algorithm}
\usepackage{array}
\usepackage[caption=false,font=small]{subfig}
\usepackage{stfloats}
\usepackage{url}
\usepackage{verbatim}
\usepackage{booktabs}
\usepackage{tikz}
\usepackage{balance}
\usetikzlibrary{arrows,automata,matrix,fit,positioning,calc}
\usepackage{pgfplots}
\usepackage{subcaption}
\usepackage{hyperref}
\usepackage[nolist,printonlyused]{acronym}
\usepackage[capitalize]{cleveref}
\usepackage[normalem]{ulem}
\pgfplotsset{compat=1.18} 

\usepackage{multirow}
\usepackage{makecell}
\usepackage[table]{xcolor}
\definecolor{headerblue}{HTML}{CFDCE9}
\definecolor{managementblue}{HTML}{EEF4FA}
\definecolor{functiongreen}{HTML}{EEF7EE}
\definecolor{methodgray}{HTML}{F2F2F2}
\definecolor{rulegray}{HTML}{666666}
\arrayrulecolor{rulegray}

\usepackage[most]{tcolorbox}
\newtcolorbox{quoteBox}{
  before skip=5pt,
  after skip=5pt,
  colback=managementblue!75,
  colframe=headerblue!100,
  left=6pt,
  right=6pt,
  top=1pt,
  bottom=1pt,
  sharp corners,
  borderline west={2pt}{0pt}{headerblue!100}
}

\begin{document}


\def\ArXivVersionCopyright{0}   

\def\IEEEVersion{1}     

\def\mycmdextendedpaper{0} 

\title{Agentic AI for Intent Management and Control in Autonomous Communication Networks}

\author{Burak Demirel, Yu Wang and Pablo Soldati\\ \IEEEmembership{Ericsson AB, Stockholm, Sweden,
email:\{name.surname\}@ericsson.com}
\if\ArXivVersionCopyright1
\thanks{\copyright~20xx IEEE. Personal use of this material is permitted. Permission from IEEE must be obtained for all other uses, in any current or future 
media, including reprinting/republishing this material for advertising or promotional purposes, creating new collective works, for resale or redistribution to servers or lists, or reuse of any copyrighted component of this work in other works.}
\else
\fi}

\markboth{Accepted for publication in the IEEE Communications Magazine}%
{Shell \MakeLowercase{\textit{et al.}}: A Sample Article Using IEEEtran.cls for IEEE Journals}


\maketitle

\begin{abstract}
Artificial intelligence (AI)-driven automation will help future mobile networks translate high-level network intents into concrete network optimization or control actions. However, existing approaches often focus on intent management and orchestration, leaving the adaptation of underlying network optimization and control functions largely unaddressed. This article proposes a triadic agentic AI framework that co-designs intent management and intent fulfillment through three agents: an interpreter, powered by a language model, that translates and refines intents into machine-readable optimization templates; an optimizer that adapts controller policies to intent specifications; and a controller that performs fast decision-making. We discuss the design space of each agent and how the workflow can be embedded across network architectural domains. Finally, we develop a concrete realization combining a language-model interpreter with Bayesian optimization for the optimizer and multi-objective reinforcement learning for the controller, and evaluate it in a radio resource management case study for differentiated connectivity. Results demonstrate how the agentic AI framework can connect high-level intents with runtime network control, showing adaptability of intent specifications and control policies to service requirements and changing network conditions.
\end{abstract}


\bstctlcite{IEEEexample:BSTcontrol} 


\acrodefplural{MDP}[MDPs]{Markov decision processes}
\acrodefplural{RTG}[RTGs]{returns-to-go}

\begin{acronym}
  \acro{AI}{artificial intelligence}
  \acro{ACK}{positive acknowledgment}
  \acro{BLER}{block error rate}
  \acro{BO}{Bayesian optimization}
  \acro{IMF}{intent management function}
  \acro{IML}{intent management loop}
  \acro{KPI}{key performance indicator}
  \acro{LA}{link adaptation}
  \acro{LM}{language model}
  \acro{MORL}{multi-objective reinforcement learning}
  \acro{OI}{operational intent}
  \acro{OLLA}{outer-loop link adaptation}
  \acro{O-RAN}{Open RAN}
  \acro{OTM}{optimization template model}
  \acro{QoE}{quality of experience}
  \acro{RAN}{radio access network}
  \acro{RAN NF}{RAN network function}
  \acro{RL}{reinforcement learning}
  \acro{RRM}{radio resource management}
  \acro{SI}{service intent}
  \acro{3GPP}{3$^\text{rd}$~Generation Partnership Project}
  \acro{4G}{4$^\text{th}$~Generation}
  \acro{5G}{5$^\text{th}$~Generation}
  \acro{6G}{6$^\text{th}$~Generation}
\end{acronym}

\acresetall 

\AddToHookNext{shipout/foreground}{%
  \put(49bp,\dimexpr-\paperheight+14bp\relax){%
    \parbox[b]{\dimexpr\paperwidth-98bp\relax}{%
      \normalfont\fontsize{8}{9}\selectfont\raggedright
      \textcopyright~2026 IEEE. Personal use of this material is permitted.
      Permission from IEEE must be obtained for all other uses, in any
      current or future media, including reprinting/republishing this
      material for advertising or promotional purposes, creating new
      collective works, for resale or redistribution to servers or lists,
      or reuse of any copyrighted component of this work in other works.\par
    }%
  }%
} 

\section{Introduction}\label{sec:1}

The next generation of \acp{RAN} will be increasingly autonomous, with management and optimization guided by high-level \emph{network intents}~\cite{TMF:21}. A network intent is a declarative directive that specifies \emph{what} the network should achieve, not \emph{how}. For example, an operator may state \emph{``maximize user coverage while minimizing energy consumption,''} leaving the network to select concrete actions—adjusting antenna tilt for coverage or deactivating carriers to save energy.

Converting intents into actions requires reasoning across multiple abstraction layers: from natural-language specifications, through optimization formalisms, down to control policies executed by \ac{RAN} functions. This conversion is challenging because intents may span multiple objectives, constraints, and network domains, while the \ac{RAN} operates under dynamic conditions and strict timing requirements. Static heuristics, rule-based orchestration, and narrowly scoped \ac{AI} models cannot bridge this gap. Bridging it requires \ac{AI} systems that interpret intents, reason over constraints, and translate goals into real-time decision-making policies.

Agentic \ac{AI} offers a promising paradigm for building such systems. Unlike monolithic models that map inputs directly to outputs, agentic \ac{AI} organizes intelligence as specialized agents that collaborate through explicit workflows: interpreting objectives, deriving sub-goals, invoking tools, and refining actions iteratively~\cite{Hughes:25}. Central to many agentic systems are generative \acp{LM}. They provide the semantic interface for interpreting declarative goals, reasoning over task context, and invoking external tools through function calls~\cite{Hughes:25}. By combining persistent-goal reasoning, tool invocation, and memory, agentic \ac{AI} can integrate planning and execution in recursive reasoning-action loops~\cite{Yao:23ReAct}, adapt over long horizons through accumulated skills and memory, and support collective operation through multi-agent communication~\cite{Wu:23AutoGen}.

Recent work has applied agentic \ac{AI} to intent-driven networking, showing that \ac{LM}-based agents can interpret and decompose intents and introduce orchestration mechanisms through structured workflows~\cite{Mekrache:24,QAJ+:25,Navidan:26}. Existing work, however, mainly focuses on high-level network and service orchestration, while largely assuming existing \ac{RAN} functions~can~handle the resulting intents. The co-design of agentic intent management with the underlying \ac{RAN} optimization and control functions, enabling policies to adapt to both intent objectives and changing network conditions, remains largely unexplored.

This paper introduces an agentic AI blueprint~that co-designs intent management and intent fulfillment through three agents: an interpreter, an optimizer, and a controller. The interpreter manages intent specifications and their refinement; the optimizer adapts the policy of the controller to intent objectives and constraints; and the controller performs runtime decision-making. Thus, the framework separates slow supervisory intent reasoning and policy adaptation from faster \ac{RAN} control while maintaining closed-loop feedback across these timescales. We discuss the design space of each agent, including LM-based interpretation and alternative methods for optimization and control. We further show how the triadic workflow can be embedded across \ac{RAN} architectural domains and applied to various network optimization and control problems. We then develop a concrete realization combining an \ac{LM}-based interpreter with \ac{BO} and \ac{MORL} to enable~policy~adaptation to intents. We evaluate this design in a~\ac{RRM} case study, demonstrating adaptation of both intent specifications and control policies to connectivity service requirements and changing network conditions.

The paper is organized as follows:~\cref{sec:2} introduces intent management in \ac{RAN}, and~\cref{sec:3} presents our agentic \ac{AI} workflow for intent-driven \ac{RAN} management and control. Sections~\ref{sec:4} and~\ref{sec:5} discuss the design alternatives for each agent, whereas~\cref{sec6:agentic_architecture} presents the framework integration across the \ac{RAN} architecture. \Cref{sec7:experiment} presents numerical results and \cref{sec:8} concludes the paper.

\section{Intent-Driven Networking} \label{sec:2}

\begin{figure*}[t!]
\centering
\includegraphics[width=0.7\textwidth]{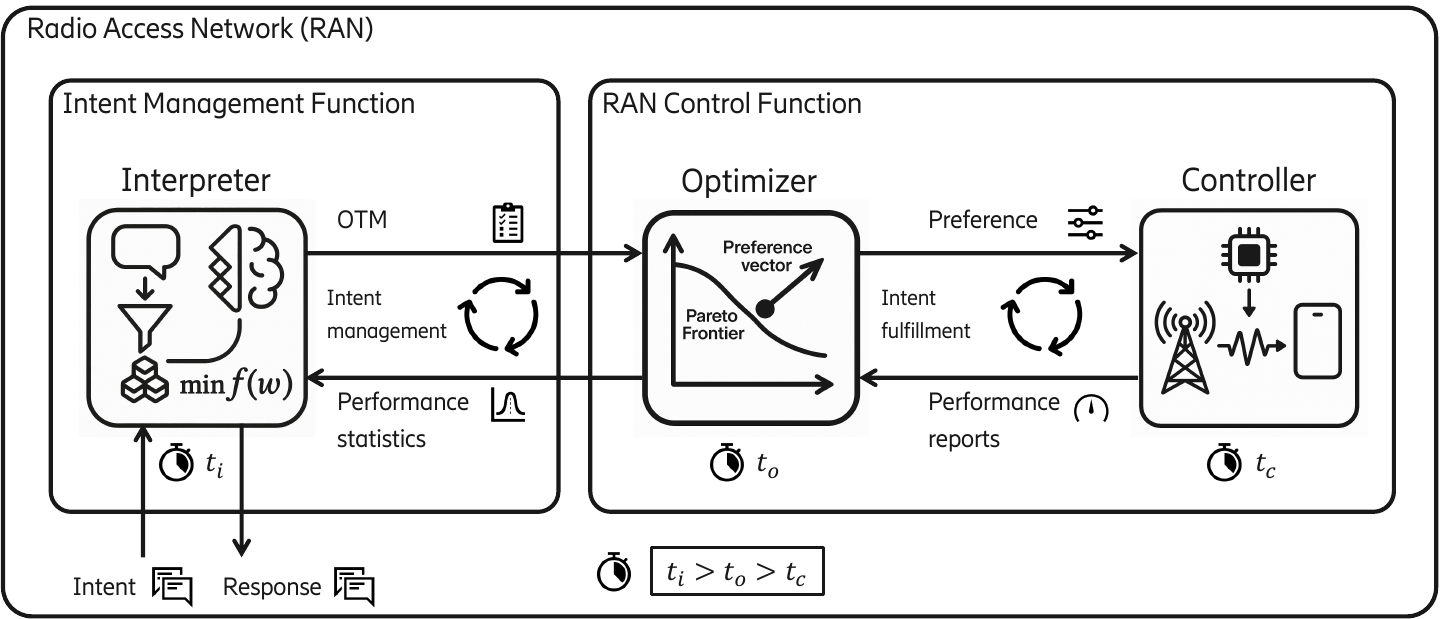}
\caption{Agentic \ac{AI} workflow for translating declarative network intents into network configurations or \ac{RAN} control actions.}
\label{fig:agentic_AI_workflow}
\end{figure*}

Network intents provide an abstraction for specifying the desired objectives, requirements, and constraints of a network service or management workflow. As they traverse network domains, intents may be expressed in declarative forms or refined into structured schemas relevant to a particular domain or network function. This enables a wide range of applications ranging from assurance, fault management, service provisioning and configuration at business, core network and service domains, to network planning, resource optimization, and \ac{RRM} for service delivery at lower \ac{RAN} domains.

While~\cite{Wang:26} provides a comprehensive survey of this field, we focus our discourse on intent-driven \ac{RAN} optimization and control and its framing into a broader architectural context. We distinguish between two intent classes relevant to this sub-area:
\begin{itemize}
\item \textbf{\Acp{OI}} specify operational objectives and requirements, such as energy, coverage, or capacity, typically related to a network area or resource pools.
\item \textbf{\Acp{SI}} specify desired end-to-end characteristics of connectivity services, such as data rate, latency, reliability and more.
\end{itemize}
\acp{OI} govern non-real-time network planning and optimization processes, typically operating over minutes to hours, and affect site-level configurations such as spectrum allocation, antenna settings, and frequency-layer activation. \acp{SI} are enforced by low-level \ac{RRM} mechanisms, such as scheduling and \ac{LA}, operating at real-time (sub-millisecond) or near-real-time (tens of milliseconds) timescales. Because the primary role of \acp{RAN} is to provide connectivity, \acp{OI} must not compromise \acp{SI}. An energy-efficiency objective, for instance, should not deactivate carriers in a way that disrupts service.

An end-to-end intent may therefore comprise multiple sub-intents with potentially overlapping or conflicting scopes and require coordinated actions across network domains. To coordinate decisions and distribute responsibilities across domains, the TM Forum structures intent orchestration as a hierarchy of \acp{IMF}, each refining intents into domain-specific sub-intents and delegating them to other domains when needed~\cite{TMF:21}. Realizing this vision requires consistent intent interpretation, feasible decomposition, and closed-loop coordination across timescales. These requirements motivate the agentic workflow introduced next.

\section{Agentic AI Blueprint} \label{sec:3}

Historically, an \emph{agent} is an autonomous entity that perceives its environment and acts toward a design objective, with or without \ac{AI}. Under this definition, \ac{RAN} systems can be modeled as distributed multi-agent systems, with agents representing control functions or network nodes at different abstraction levels. However, \ac{RAN} systems are largely non-agentic in their behavior, relying on predefined heuristics and coordination logic rather than adaptive, goal-driven reasoning and planning.

\Cref{fig:agentic_AI_workflow} presents a blueprint for embedding agentic \ac{AI} in \acp{RAN} for intent-driven management and control. At its core are three agents: an \emph{interpreter}, an \emph{optimizer}, and a \emph{controller}. Their interactions form an agentic workflow comprising two loops: an \emph{intent management loop}, executed by the interpreter--optimizer pair, and an \emph{intent fulfillment loop}, executed by the optimizer--controller pair. These loops operate at different timescales, forming a two-loop, multi-timescale control architecture that is loosely analogous to Kahneman's dual-process model~\cite{Kahneman:11book}, with a slower deliberative outer loop resembling System~2 and a faster reactive inner loop resembling System~1.

The \emph{interpreter} decomposes and encodes intents into structured \acp{OTM}, invokes the appropriate \ac{RAN} functions, and refines intents using network observations and fulfillment feedback. The \emph{optimizer} adapts the controller's policy to align it with the intent specifications and summarizes fulfillment feedback for the interpreter. The \emph{controller} performs runtime decision-making, collects observations, and reports performance metrics to the optimizer.

\Cref{sec6:agentic_architecture} extends the discussion to \ac{RAN} architecture domains in two dimensions. Horizontally, within a \ac{RAN} domain, an interpreter can oversee multiple optimizer--controller pairs integrated in different \ac{RAN} functions. Vertically, we integrate the workflow in different \ac{RAN} domains to support operations ranging from fast \ac{RRM} to slower network management.

\subsection{Timescale Separation}

The workflow distributes responsibilities across three timescales, as exemplified in~\cref{tab:agent_mapping}. The controller defines the system's reference timescale, denoted by $t_{c}$. Since the controller is associated with a specific \ac{RAN} function, it inherits that function's native latency budget, ranging from sub-millisecond timescales for real-time \ac{RRM} tasks, such as link adaptation, to minutes or hours for non-real-time network management functions, such as cell shaping. The optimizer adapts the controller's policy at a deliberately slower timescale $t_{o}$. This timescale ranges from hundreds of milliseconds to seconds for \ac{RRM} functions and up to hours for \ac{RAN} management tasks, separating policy adaptation from the control loop. The interpreter operates at the slowest supervisory cadence, denoted by $t_{i}$, spanning seconds to minutes for \ac{RRM} supervision and up to hours for network-level management.

Separating long-term, non-real-time reasoning and intermediate policy adaptation from real-time control ensures that (i) interpreter decisions remain non-time-critical, and (ii) latency-sensitive operations are confined to the controller.

\section{Language-Guided Network Management}\label{sec:4}

\begin{table*}[t]
    \centering
    \caption{Mapping of agent roles to \ac{RAN} functionalities,
    timescales, and applicable methodologies across RAN domains.}
    \label{tab:agent_mapping}

    \setlength{\tabcolsep}{0pt}
    \renewcommand{\arraystretch}{1.15}

    \begin{tabular}{
        @{}>{\raggedright\arraybackslash}p{3.5cm}
           >{\raggedright\arraybackslash}p{2.5cm}
           >{\raggedright\arraybackslash}p{3.5cm}
           >{\raggedright\arraybackslash}p{4.5cm}
           >{\raggedright\arraybackslash}p{4.0cm}@{}
    }

    \rowcolor{headerblue}
    \multicolumn{2}{@{}>{\centering\arraybackslash}p{6.0cm}@{}}
        {\textbf{Network domain}}
    &
    \multicolumn{1}{@{}>{\centering\arraybackslash}p{3.5cm}@{}}
        {\textbf{Interpreter}}
    &
    \multicolumn{1}{@{}>{\centering\arraybackslash}p{4.5cm}@{}}
        {\textbf{Optimizer}}
    &
    \multicolumn{1}{@{}>{\centering\arraybackslash}p{4.0cm}@{}}
        {\textbf{Controller}}
    \\
    \hline

    \rowcolor{managementblue}
    &
    \textbf{Timescales}
    &
    Non-real-time 
    [$10^{3}$--$10^{5}$\,s]
    &
    Non-real-time 
    [$10^{2}$--$10^{4}$\,s]
    &
    Non-real-time 
    [$>10^{2}$\,s]
    \\

    \rowcolor{managementblue}
    \multirow{-2}{=}{\textbf{RAN Management}}
    &
    \textbf{Functionalities}
    &
    IMF\newline
    Orchestration and management \newline
    Standalone
    &
    \multicolumn{2}{
        @{}>{\raggedright\arraybackslash}p{8.5cm}@{}
    }{
        Network planning (resource configuration,
        network slicing, etc.)\par
        Network optimization (cell shaping,
        energy optimization, etc.)
    }
    \\
    \hline

    \rowcolor{functiongreen}
    &
    \textbf{Timescales}
    &
    Non-real-time 
    [$>10^{0}$\,s]
    &
    Near-real-time 
    [$10^{-1}$--$10^{0}$\,s]
    &
    Real-time 
    [$10^{-4}$--$10^{-2}$\,s]
    \\

    \rowcolor{functiongreen}
    \multirow{-2}{=}{\textbf{RAN Network Function}}
    &
    \textbf{Functionalities}
    &
    IMF\newline
    Standalone
    &
    \multicolumn{2}{
        @{}>{\raggedright\arraybackslash}p{8.5cm}@{}
    }{
        Medium access control (link adaptation,
        scheduling, etc.)\par
        Network self-organization (load balancing,
        mobility optimization, etc.)
    }
    \\
    \hline

    \rowcolor{methodgray}
    \multicolumn{2}{
        @{}>{\raggedright\arraybackslash}p{6.0cm}@{}
    }{
        \textbf{Applicable Methodologies}
    }
    &
    LM-based methods\newline
    Semantic parsing\newline
    Heuristics
    &
    Optimization theory\newline
    Stochastic approximation\newline
    Constrained RL\newline
    Evolutionary algorithms\newline
    Bayesian optimization
    &
    Heuristics\newline
    Optimization and control theory\newline
    Reward-free RL\newline
    Reward- or goal-conditioned RL\newline
    Multi-objective RL
    \\
    \hline

    \end{tabular}
\end{table*}

The interpreter is a supervisory cognitive agent responsible for translating intents into executable workflows for \ac{RAN} control and management. Interpreting intents extends beyond simple language parsing. The interpreter must (i) reason over the capabilities, operational scope, and constraints of available \ac{RAN} functions; (ii) determine feasible decompositions and valid optimization objectives; (iii) invoke the appropriate \ac{RAN} functions for intent handling; and (iv) coordinate intent refinement across network domains when required.

We therefore adopt an \ac{LM}-guided design, which allows the interpreter to accept intents in both declarative and structured form. Upon receiving an intent, the interpreter first analyzes its semantic scope, objectives, constraints, and service requirements to determine the relevant \ac{RAN} domains and control functions. It then evaluates whether the intent can be directly fulfilled or must be decomposed into smaller sub-intents associated with different functions, timescales, or network domains. Once the intent is structured and the appropriate \ac{RAN} function handler identified, the interpreter encodes it as an \ac{OTM} instance for downstream optimization and control.

The proposed \ac{OTM} schema organizes intents into machine-readable blocks that separate objectives, constraints, requirements, and metadata, thereby enabling consistent communication across interpreters and downstream \ac{RAN} functions. The metadata block supports intent management and provenance tracing by recording origin, ownership, scope, priority, validity period, decomposition status, and many more, allowing the interpreter to distinguish external, delegated, or refined intents and trace them across network domains.

We provide the interpreter with domain awareness by exposing \ac{RAN} functions as callable tools. Inspired by the Model Context Protocol, \ac{RAN} functions can advertise machine-readable descriptors specifying operational scope, supported parameters, controllable performance metrics, and timescales. Using these descriptors, the interpreter can identify suitable intent handlers and formulate proper sub-intents.

Finally, the interpreter can perform supervisory reasoning over network observations and fulfillment feedback. When changing network conditions or conflicting objectives make an intent infeasible, it may refine requirements or adapt trade-offs, subject to the authority and bounds specified in the intent. This enables local adaptation where needed without assuming unrestricted authority to modify upstream requirements.

\section{Intent Fulfillment Loop}\label{sec:5}

\subsection{Optimizer Agent}

The optimizer performs three tasks: (i) decoding the \ac{OTM} provided by the interpreter and formulating the associated optimization problem, (ii) recursively solving this problem to align the controller's policy with the specified intent, and (iii) coordinating feedback between the intent management and fulfillment loops. Upon receiving an \ac{OTM}, the optimizer formulates the intent as a constrained optimization problem:
\begin{equation}
    \begin{aligned}
        \underset{\boldsymbol{\omega} \in \Omega}{\text{minimize}} \quad & f(\boldsymbol{\omega}) \\
        \text{subject to} \quad & g_{i}(\boldsymbol{\omega}) \leq b_i, \quad i = 1, \dots, p,
    \end{aligned}
    \label{eq:optimization}
\end{equation}
where \( f(\boldsymbol{\omega}) \) quantifies the intent objectives, such as energy cost, latency, or throughput. The decision variables \( \boldsymbol{\omega}\in\Omega \subseteq \mathbb{R}^m \) represent hyperparameters that tune the controller's policy. The constraints \( g_i(\boldsymbol{\omega}) \leq b_i \) capture operational limitations (e.g., bandwidth, latency, or power) and service requirements. Because the objective and constraints are often non-convex, the solution landscape may contain multiple local optima. While~\eqref{eq:optimization} describes a local optimization problem within a \ac{RAN} domain, its scope can encompass multiple downstream RAN control or network optimization functions.

The optimizer can be designed using a wide range of methodologies, each offering different trade-offs between adaptability, computational cost, and robustness. Classical gradient-based and gradient-free methods are effective when the optimization problem is well characterized and evaluations are inexpensive. Stochastic approximation methods, such as stochastic gradient descent and simultaneous perturbation stochastic approximation, can enable lightweight online adaptation from noisy measurements, like observations in time-varying network conditions. Bayesian optimization offers a sample-efficient alternative for expensive, noisy, and non-convex black-box problems, making it suitable for optimizing controller configurations when evaluations are costly or gradient information is unavailable. Evolutionary algorithms, such as genetic algorithms and particle swarm optimization, provide another gradient-free approach using population-based search to explore complex, non-differentiable, or multi-modal optimization landscapes, although their computational cost can limit their use to offline or slow adaptation.

\subsection{Controller Agent}\label{sec:5_controller}

The controller handles intents through runtime decision-making. A key design requirement is runtime \emph{policy adaptability}: the ability to adjust its behavior to the objectives and trade-offs specified by different intents under varying network conditions, without manual reconfiguration or retraining. For example, a resource scheduler may need to dynamically shift its strategy among throughput, latency, and reliability depending on the requirements of different connectivity services.

Run-time policy adaptability can be achieved with a wide range of methods. Rule-based heuristics, classical optimization and control remain viable when the system dynamics and performance trade-offs are well understood. These approaches offer interpretability, predictable behavior, and low computational overhead. For stochastic environments with high-dimensional state spaces, data-driven methods such as \ac{RL} become attractive. Conventional \ac{RL}, however, typically optimizes a fixed reward function, requiring separate models to accommodate different objectives.

To overcome the fixed-reward limitation of conventional \ac{RL}, recent approaches seek to generalize policies across multiple reward specifications. These include reward-free \ac{RL}~\cite{JKS:20reward_free}, which explores the environment without predefined rewards; multi-reward approaches, which seek optimal policies over a predefined set of reward functions~\cite{russo2024multireward}; and reward- or goal-conditioned policy learning~\cite{Emmons2022RvS}. While promising, these approaches cannot directly optimize trade-offs among multiple objectives and rely on assumptions that can be challenging to satisfy in \acp{RAN}, such as extensive exploration over many or unknown rewards and reliable estimates of future values.

Among these approaches, \ac{MORL}~\cite{YSN:19} directly addresses the challenge of modeling trade-offs among multiple objectives. \ac{MORL} considers a vector of rewards expressing different performance objectives and learns a single policy that approximates the Pareto frontier of achievable reward trade-offs. The training process jointly explores the state-action space and a preference space representing the importance of each reward component. The \ac{MORL} exploration efficiency in \acp{RAN} can be improved by distributing it across \ac{RAN} nodes, with different nodes exploring different regions of the preference space. At runtime, the policy is conditioned on a relative importance assigned to each objective to achieve a desired behavior.

\subsection{Intent Fulfillment Loop Design - An Example}\label{sec:5_fulfilment_loop}

The appropriate design choice for the optimizer-controller agents is not unique. It depends on the properties of the optimization problem and the operational requirements of the RAN functionality they replace, including the adaptation timescale, evaluation cost, computational complexity, non-stationarity, and safe constraint handling.

One realization we propose to achieve runtime policy adaptability that suits several RAN functions (e.g., scheduling, link adaptation, and cell-shaping) combines a \ac{BO}-based optimizer with a \ac{MORL} controller. The optimizer uses controller feedback to recursively refine~\eqref{eq:optimization} and derive the controller hyperparameters $\boldsymbol{\omega}^{\star}$, here representing the preference among intent objectives under observed network conditions. Conditioned on $\boldsymbol{\omega}^{\star}$, the \ac{MORL} policy maps the current \ac{RAN} function state to real-time actions, enabling closed-loop adaptation toward Pareto-efficient operating points aligned with the intent. \Cref{sec7:experiment} demonstrates this approach in an example.

\section{Agentic AI in the RAN Architecture}\label{sec6:agentic_architecture}

\begin{figure*}[t!]
    \centering
    \includegraphics[width=0.92\textwidth]{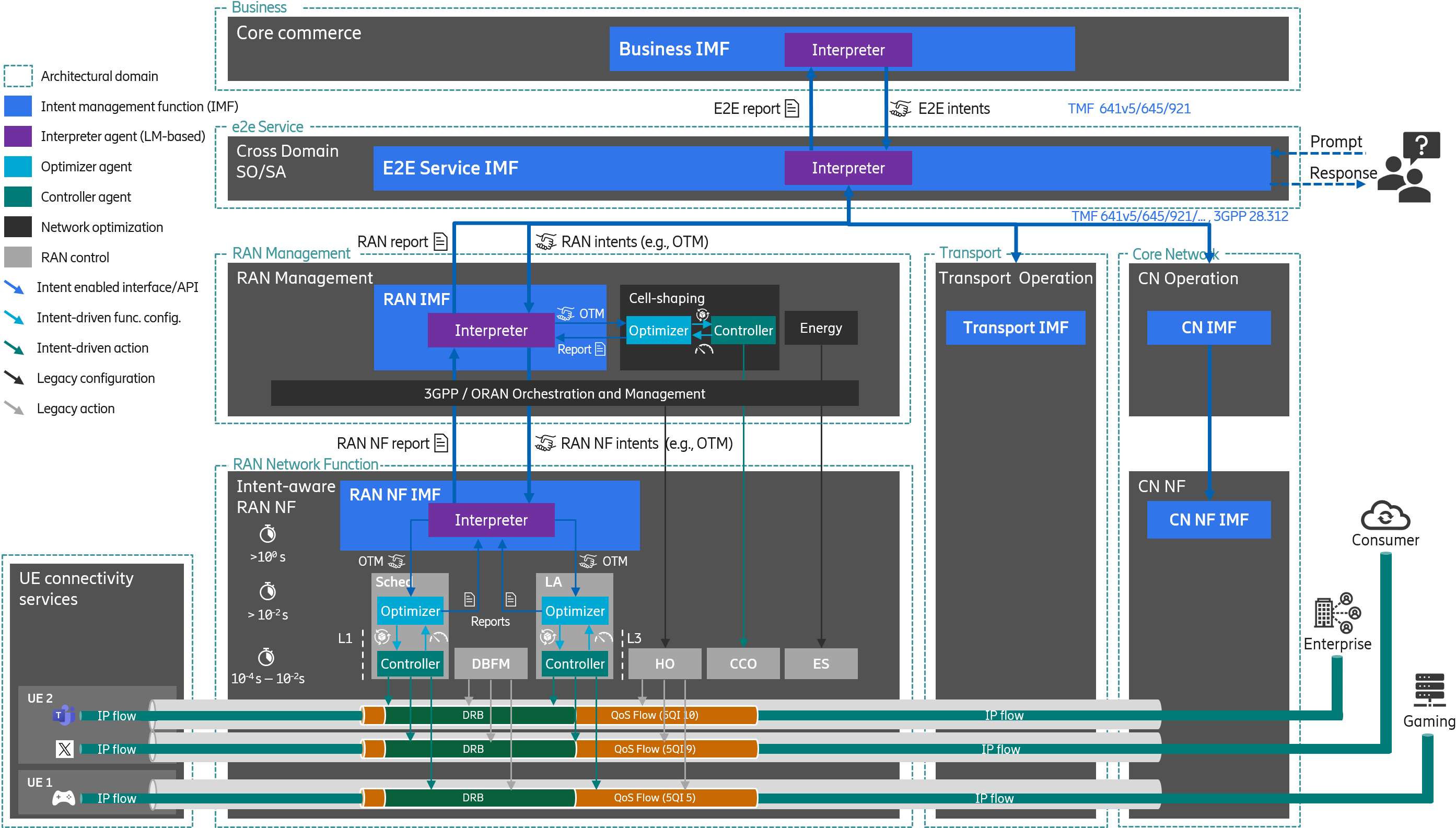}
    \caption{An agentic \ac{AI} architecture for intent management compatible with the layered orchestration defined by the TM Forum.}
    \label{fig:agentic_AI_architecture}
\end{figure*}

\subsection{Agentic AI Architecture}

We presented the agentic \ac{AI} workflow in a domain-agnostic manner. We now consider its integration across a RAN architecture. \Cref{tab:agent_mapping} exemplifies agent roles, timescales, and possible functional mappings for two domains, and \Cref{fig:agentic_AI_architecture} extends the analysis to a broader architectural~context. While~we discuss cross-domain intent coordination, implementing and evaluating optimization across domains~remains future work.
 
The interpreter provides the cognitive interface to downstream \ac{RAN} control functions or entire network domains, while optimizer--controller pairs provide a paradigm for redesigning network optimization and \ac{RAN} control functions. An interpreter may therefore be hosted by management and coordination functions at different levels, from business and service assurance to non-real-time \ac{RAN} network orchestration or service management, and near-real-time management in the lowest \ac{RAN NF} domain.

For ease of description and compatibility with existing \ac{RAN} architectures, \Cref{fig:agentic_AI_architecture} uses the hierarchy of \acp{IMF}~\cite{TMF:21} as an architectural baseline and maps the interpreter within the \ac{IMF} functional scope. Thus, with this choice, a network domain may require at most one \ac{LM}-based interpreter agent. While higher-domain interpreters may share a common cloud infrastructure, deployment in lower domains requires consideration of available compute and memory resources. These aspects and alternative architectures are discussed in \Cref{secVIC:extensions}.

The interpreter agent enables the \ac{IMF} to interpret, structure, and operationalize intents using domain awareness by exposing downstream \ac{RAN} functions or underlying network domains as functional tools. The interpreter can then establish two types of \acp{IML}:
\begin{itemize}
    \item \textbf{An Intra-domain \ac{IML}} operates within a \ac{RAN} domain.
    \item \textbf{An Inter-domain \ac{IML}} operates across \ac{RAN} domains.
\end{itemize}

Within a domain, when a suitable function is identified to handle an intent, the interpreter establishes an intra-domain \ac{IML} and provides an initial \ac{OTM} to the corresponding~optimizer. The latter then initiates the intent-fulfillment loop, adapting the controller policy to the intent specifications.~When an intent cannot be handled within the current domain, the interpreter can delegate it to an interpreter in an underlying domain via an inter-domain \ac{IML}. Standardized protocols such as those defined by the TM Forum can provide the communication framework between \acp{IMF}, while agentic collaboration may require extensions inspired, for instance, by emerging agent-to-agent communication. In both cases, downstream intent handling is initiated with structured \acp{OTM}.

We illustrate these principles using examples from two domains in~\Cref{fig:agentic_AI_architecture}. In the \ac{RAN} management domain, an interpreter hosted by a \ac{RAN} \ac{IMF} owns operational intents for network-level optimization. An operator may specify an intent to balance coverage and capacity across a network area. The interpreter may select a cell-shaping function and establish an intra-domain \ac{IML} with it. This function can employ the optimizer--controller paradigm, e.g., with a preference-based \ac{MORL} controller, to adapt the desired coverage--capacity trade-off at runtime.

Furthermore, service-related intents can be delegated to an underlying interpreter hosted in a \ac{RAN NF} \ac{IMF} with~an inter-domain \ac{IML}. The latter can further decompose the received intent into service-specific sub-intents and establish intra-domain \acp{IML} with suitable \ac{RAN} functions. \Cref{fig:agentic_AI_architecture} shows two examples of the optimizer--controller paradigm applied to service-level control: resource scheduling, which can adapt resource allocation to latency or throughput requirements, and link adaptation, which can adjust transmission parameters to meet reliability or rate objectives. For instance, the scheduler may combine a rate optimizer establishing target rates for each connectivity service with a time-frequency resource allocator. These examples illustrate how the same workflow can support both network-level optimization and lower-level \ac{RAN} control.

Finally, fulfillment information propagates bottom-up: controllers report performance to their optimizers, while \acp{IMF} exchange feasibility and fulfillment information across domains. This feedback enables interpreters to refine requirements, relax constraints, or generate alternative sub-intents when necessary. An interpreter at a lower domain, for instance, may receive an \ac{OTM} delegating refinement authority by indicating which requirements or constraints can be adapted, under which conditions, and within what bounds, so that local refinement does not imply unrestricted authority over upstream owners.

\subsection{Practical Considerations and Future Extensions} \label{secVIC:extensions}

Deploying the proposed agentic workflow in present \ac{RAN} systems requires accounting for constraints posed by compute platforms tailored to communication workloads and standardized architectures. Thereby, our design considers~(a) a lightweight \ac{LM}-based interpreter, e.g., a compact Qwen, Phi, or Llama variant, dimensioned to available \ac{RAN} compute resources; and (b) an architecture anchored to existing hierarchies of \acp{IMF}. These choices facilitate integration and limit the need for dedicated AI infrastructure.

These choices, however, may be revisited in next-generation 6G \ac{RAN} systems. One approach is to introduce a dedicated AI domain in the architecture to host interpreter agents serving multiple interdependent \ac{RAN} domains and more flexible orchestration mechanisms. For example, \Cref{fig:agentic_AI_architecture} could be redrawn with an \ac{AI} domain hosting two connected interpreter agents serving the business/end-to-end service domains and the \ac{RAN} management/\ac{RAN NF} domains, respectively. This would allow centralizing these agents in a shared cloud infrastructure, enable more efficient resource utilization and support larger models. It could also facilitate operator interaction with agents and provide visibility into lower-level \ac{RAN} domains.

\begin{figure*}[t!]
    \centering
        \subfloat[Time series of service performance for two connectivity services.]{\includegraphics[width=0.48\textwidth]{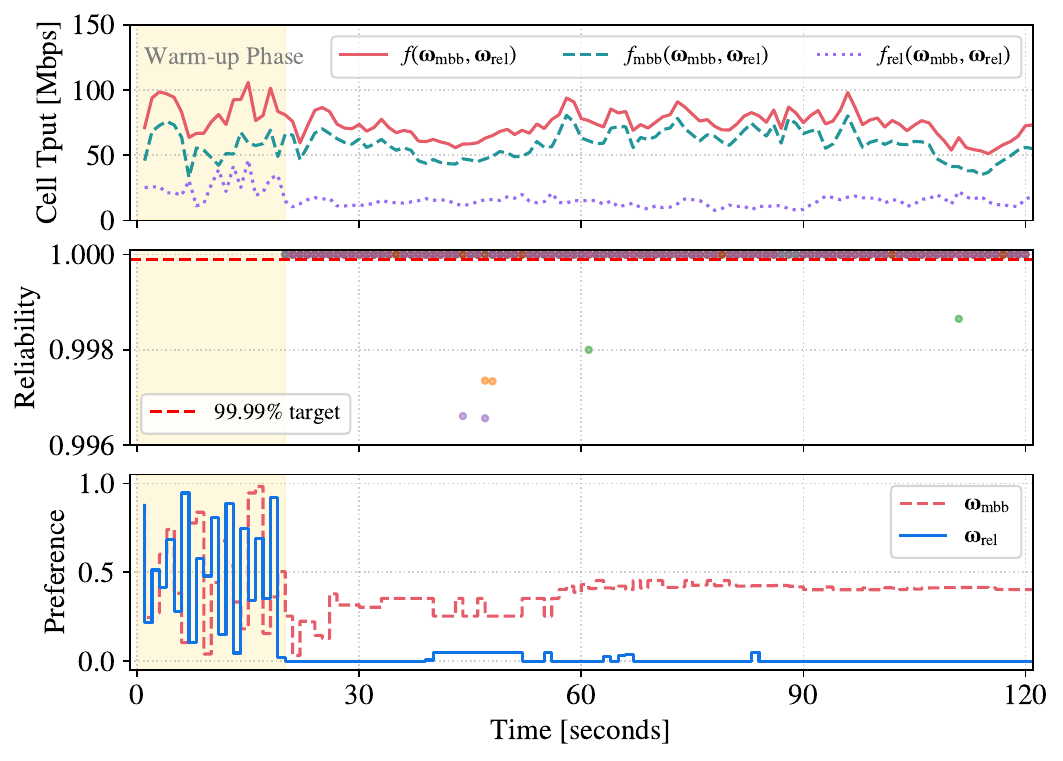}%
        \label{fig:3a}
        }
        \hfill
        \subfloat[Time series of service performance with rate requirement.]{
        \includegraphics[width=0.47\textwidth]{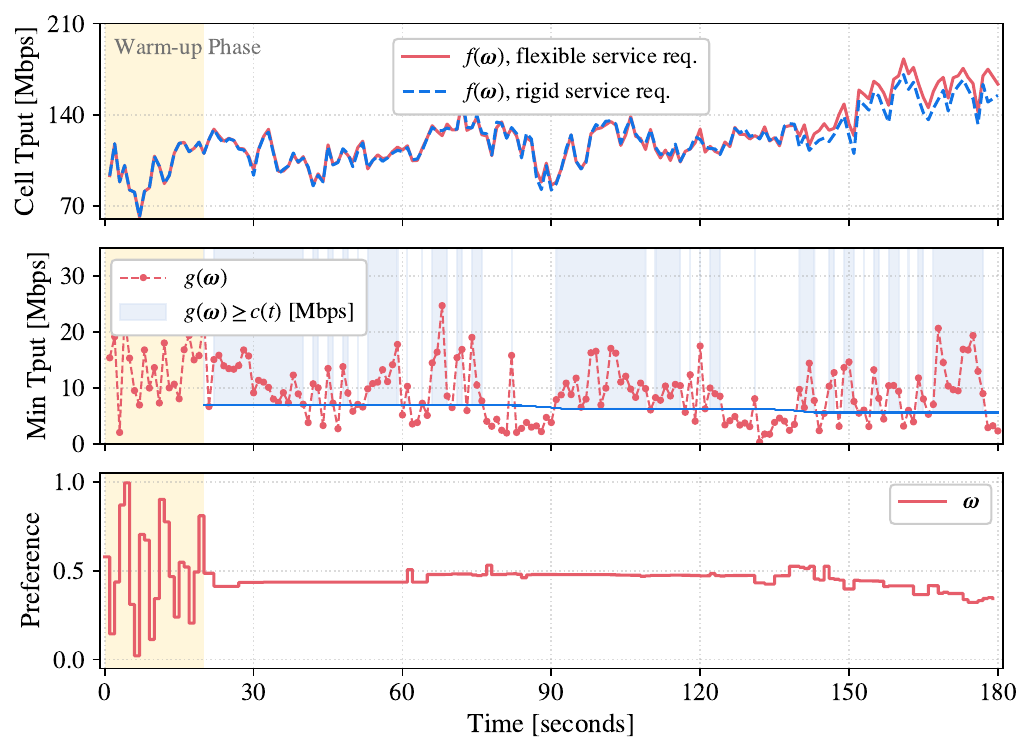}%
        \label{fig:3b}
        }
    \caption{Demonstration of the agentic system with two examples for supporting two differentiated connectivity services.}
    \label{fig:3_optimizer_controller_loop}
\end{figure*}

\section{Numerical Evaluations}\label{sec7:experiment}

We apply the agentic \ac{AI} system in~\Cref{fig:agentic_AI_workflow} to an intent-aware \ac{RRM} case study for differentiated connectivity services, combining an \ac{LM}-based interpreter agent with an intent-driven \ac{LA} function that integrates a \ac{BO}-based optimizer and a \ac{MORL} controller. Design details are available in a preprint~\cite{demirel:25}.

\subsection{Interpreter and Timescale Dimensioning} 

We design the interpreter with a Qwen3.5-9B model executed on an NVIDIA A100 GPU whose specifications are comparable to accelerators deployed in today's \ac{RAN} systems. 

On a 1,000-intent synthetic test set, the Qwen3.5-9B interpreter produced schema-valid OTMs in all cases and semantically correct \acp{OTM} in 92.0\% of cases, with a mean latency of 2.98 $\pm$ 0.92 s. An extended evaluation is available in~\cite{demirel:25}.

Applying agentic \ac{AI} to \ac{RRM} is particularly challenging due to stringent execution-latency budgets. We dimension each agent to comply with realistic system requirements. The \ac{LA} controller acts at each transmission time interval ($t_c\leq 1$\,ms). The optimizer adapts the controller policy every few hundred ms ($t_o\in[0.5,\; 1]$\,s). The interpreter operates at a supervisory timescale $t_i$ at least an order of magnitude slower than~$t_o$; its inference is event-driven, triggered by persistent intent violations, and occurs on the order of seconds. This separation decouples \ac{LM} inference from real-time control.

\subsection{Intent-Driven Link Adaptation} \label{subsec:MORL_controller_LA}

We design a \ac{MORL} \ac{LA} controller with a two-dimensional reward $\mathbf{r} = [r_1, r_2]^{\top}$ whose components capture competing objectives: $r_1$ represents the amount of information bits successfully carried by a packet, while $r_2$ is the time-frequency resource cost of each transmission. Policy adaptation is governed by a preference vector $\boldsymbol{\omega} = [\omega, 1-\omega]^{\top}$, with $\omega\in[0,\;1]$.

{\color{black}We further consider a \ac{BO}-based optimizer to adapt $\omega$ and guide the \ac{MORL} \ac{LA} controller policy to meet the service requirements under prevailing network conditions. When $\omega\approx 0$, the controller applies a conservative \ac{LA} policy, resulting in reliable transmissions at the cost of low throughput and spectral efficiency. When $\omega \approx 1$, it applies an aggressive \ac{LA} policy that maximizes spectral efficiency at the expense of retransmissions and throughput. The best throughput operating points emerge at intermediate $\omega$ that depend on network~conditions.}

\vspace{-2mm}
\subsection{Triadic Agentic Framework Validation}
\label{subsec:optimizer_controller_loop}
We evaluate the agentic workflow with a high-fidelity, event-driven system-level simulator compliant with the 3GPP 5G-A specifications. Each simulation models a heterogeneous multi-cell \ac{RAN} operating in time division duplexing mode with single-user multiple-input multiple-output transmission. We use a carrier frequency \qty{3.5}{GHz} with a {\color{black}\qty{100}{MHz} band}, and a physical layer compliant with the 5G specifications for numerology $\mu=0$. We refer to~\cite{demirel:25} for a complete description.

\Cref{fig:3a} demonstrates the triadic agentic system for intent-driven \ac{RRM} using an intent that combines two contrasting connectivity services under a common system objective:
\begin{quoteBox}
    \emph{Maximize cell throughput while serving mobile broadband users on a best-effort basis and guaranteeing 99.99\% reliability for reliability-critical traffic.}
\end{quoteBox} 
This case study is designed to demonstrate near-real-time policy adaptation, with a single MORL model, to balance the contrasting goals of best-effort and reliability-critical services.

\Cref{fig:3a} shows the agentic loop dynamics in a two-minute simulation. {\color{black}Here, reliability measures the ratio of packets correctly received over \qty{1}{s}}. Upon obtaining an \ac{OTM}, the optimizer adapts separate weights $\boldsymbol{\omega}_\mathrm{rel}$ and $\boldsymbol{\omega}_\mathrm{mbb}$ so the controller applies Pareto-optimal policies matched to each service's requirements under varying network conditions. The~optimizer steers the \ac{MORL} controller towards a conservative \ac{LA} policy with $\omega_{\mathrm{rel}}\approx 0$ for reliable services, whilst it drives the controller to a more aggressive \ac{LA} policy with $\omega_{\mathrm{mbb}}\approx 0.45$ to provide higher mean throughput to streaming~users. During the simulation, the interpreter monitors the intent fulfillment with feedback from the optimizer. Since no persistent intent violation occurs, the interpreter maintains the \ac{OTM} unchanged.

\Cref{tab:performance} compares our system with the state-of-the-art 5G \ac{LA}, based on a control loop that targets a fixed \ac{BLER}, and the \ac{RL}-based \ac{LA}  in~\cite{demirel:2026}. Unlike our approach—which adapts a single \ac{MORL} model {\color{black}in near real time} to meet different service requirements—both classical \ac{RL} and 5G \ac{LA} require separate configurations per service. {\color{black}Classical} \ac{RL} requires two distinct models with service-specific rewards; following~\cite{demirel:2026}, we set robustness parameters $\alpha=0.5$ for throughput and $\alpha=2$ for reliability. Similarly, for 5G \ac{LA} we use the target \ac{BLER} (i.e., the loss probability of the first transmission) as a conservative surrogate for reliability, selecting a typical 10\% for streaming and 1\% for reliability-critical services. The results, obtained over 100 runs, show that our system achieves substantially lower \ac{BLER} for reliability-critical services than both baselines, while attaining throughput comparable to an \ac{RL} model trained explicitly for throughput.

\Cref{fig:3b} illustrates the application of the agentic system to a different service example defined by the following intent:
\begin{quoteBox}
\emph{Maximize cell throughput and serve streaming users with a minimum average data rate of \qty{7}{Mbps} when possible.}
\end{quoteBox}
This example considers a high rate demand to stress the system, as its feasibility can vary with channel conditions, interference, load, and mobility. The purpose is not to guarantee the specified rate under all conditions, but to demonstrate how the agentic system adapts to maximize cell {\color{black}throughput} when the requirement cannot be sustainably satisfied. \Cref{fig:3b} compares two settings: (a) rigid service requirements, where the minimum-rate constraint is enforced regardless of network conditions, and (b) flexible service requirements, where the \ac{OTM} permits {\color{black}relaxation of} the service constraint. In this case, when the optimizer alerts the interpreter of a consistent constraint violation, the interpreter reasons about the violation and refines the service constraint, providing a revised \ac{OTM} as in the following example, where the minimum rate is reduced.

\begin{tcolorbox}[
    before skip=5pt,
    after skip=5pt,
    colback=managementblue!75,
    colframe=headerblue!100,
    boxrule=0.4pt,
    arc=0pt,
    left=40pt,  
    right=4pt,
    top=-4pt,    
    bottom=-4pt  
]
\begin{lstlisting}[
    basicstyle=\ttfamily\scriptsize,
    columns=fullflexible,
    keepspaces=true,
    breaklines=true,
    frame=none
]
{
  "version": "1.09",
  "objective": {
  },
  "constraints": [{
    "threshold": 4.25,
    "id": "C1"
  }],
  "metadata": {
    "origin": "RAN_NF_IMF",
    "owner": "RAN_NF_IMF",
    "handler": "link_adaptation",
    "episode": "alert_009",
    "timestamp": "2026-04-22T10:20:00Z",
    "timescale": "10s_window",
    "revised": true,
    "revised_by": "RAN_NF_IMF"
  }
}
\end{lstlisting}
\end{tcolorbox}

The results show that by reducing the rate requirement, the interpreter enables the optimizer to guide the \ac{MORL} controller toward a less aggressive \ac{LA} policy for streaming users with poor channel conditions, protecting overall cell throughput. {\color{black}In this particular case, cell throughput improved by 7.0\%.}

\section{Conclusions} \label{sec:8}

We presented an agentic \ac{AI} blueprint for intent-driven \ac{RAN} management and control that co-designs intent management and intent fulfillment through three cooperating agents: an interpreter, an optimizer, and a controller. The interpreter manages and refines intent, the optimizer adapts controller policies to such specifications, and the controller {\color{black}makes runtime} decisions. Their interaction separates {\color{black}slower} supervisory reasoning and policy adaptation from faster \ac{RAN} control while maintaining feedback loops over multiple timescales.

We discussed the design space of each agent, including \ac{LM}-based interpretation and alternative optimization and control methods, and showed how the workflow can be embedded across RAN architectural domains and applied to different optimization and control problems. We further developed a concrete realization and evaluated it through an intent-aware \ac{RRM} case study, demonstrating adaptation to connectivity-service objectives and changing network conditions.

Future work will extend the framework to multi-domain orchestration, including intent decomposition and coordination across RAN domains, and will evaluate these extensions with additional RAN management use cases.

\begin{table}[t]
    \centering
    \begin{tabular}{lcccccc}
        \toprule
        & \multicolumn{3}{c}{BLER [\%] reliable service} & \multicolumn{3}{c}{Rate [Mbps] streaming} \\
        \cmidrule(lr){2-4} \cmidrule(lr){5-7}
        Algorithm & 5th & 50th & 90th & 5th & 50th & 90th \\
        \midrule
        Agentic \ac{LA}     & $\mathbf{~0.00}$ & $\mathbf{~0.39}$ & $\mathbf{~1.79}$ & $15.76$ & $\mathbf{23.21}$ & $\mathbf{29.23}$ \\
        \ac{RL} $\alpha=0.5$& $~6.34$ & $12.86$ & $17.26$ & $\mathbf{15.95}$ & $23.01$ & $28.26$ \\
        \ac{RL} $\alpha=2.0$& $~2.65$ & $~6.29$ & $~9.08$ & $14.35$ & $21.75$ & $27.45$ \\
        5G \ac{LA} $1\%$        & $~0.07$ & $~0.91$ & $~2.82$ & $~8.98$ & $13.38$ & $16.94$ \\
        5G \ac{LA} $10\%$       & $~6.98$ & $~9.36$ & $11.10$ & $13.60$ & $19.40$ & $24.24$ \\
        \bottomrule
    \end{tabular}
    \caption{Performance comparison across algorithms.}
    \label{tab:performance}
\end{table}

\bibliographystyle{IEEEtran}
\balance

\bibliography{references_very_short.bib}

\end{document}